\documentclass[runningheads]{llncs}

\usepackage{graphicx} 
\usepackage{epsfig} 
\usepackage{mathptmx} 
\usepackage{times} 
\usepackage{amsmath} 
\usepackage{amssymb}  
\usepackage{textcomp}
\usepackage{caption}
\usepackage{url}
\usepackage{hhline}
\usepackage{multirow}
\usepackage{colortbl}
\usepackage{xcolor}
\usepackage[utf8]{inputenc}

\title{\LARGE \bf
AIBA: An AI Model for Behavior Arbitration in Autonomous Driving
}

\author{B. Tr\u{a}snea\inst{1,2}, C. Pozna\inst{1,3} and Sorin M. Grigorescu\inst{1,2}
}

\institute{Robotics, Vision and Control Lab, Transilvania University of Brasov, Romania 
\email{\{bogdan.trasnea, cp, s.grigorescu\}@unitbv.ro}
\and Department of Artificial Intelligence, Elektrobit Automotive
\and Department of Informatics, Széchenyi István University of Gyõr, Hungary}

\begin{document}

\maketitle
\thispagestyle{empty}
\pagestyle{empty}

\begin{abstract}

Driving in dynamically changing traffic is a highly challenging task for autonomous vehicles, especially in crowded urban roadways. The Artificial Intelligence (AI) system of a driverless car must be able to arbitrate between different driving strategies in order to properly plan the car’s path, based on an understandable traffic scene model. In this paper, an AI behavior arbitration algorithm for Autonomous Driving (AD) is proposed. The method, coined AIBA (AI Behavior Arbitration), has been developed in two stages: (i) human driving scene description and understanding and (ii) formal modelling. The description of the scene is achieved by mimicking a human cognition model, while the modelling part is based on a formal representation which approximates the human driver understanding process. The advantage of the formal representation is that the functional safety of the system can be analytically inferred. The performance of the algorithm has been evaluated in Virtual Test Drive (VTD), a comprehensive traffic simulator, and in GridSim, a vehicle kinematics engine for prototypes.

\end{abstract}


\section{Introduction and Related Work}

Autonomous Vehicles (AVs) are robotic systems that can guide themselves without human operators. Such vehicles are equipped with \textit{Artificial Intelligence} (AI) components and are expected to change dramatically the future of mobility, bringing a variety of benefits into everyday life, such as making driving easier, improving the capacity of road networks and reducing vehicle-related accidents.

Most likely, due to the lack of safety guarantees and legislation, as well as the missing scalability of \textit{Autonomous Driving} (AD) systems, fully autonomous vehicles will not travel the streets in the near future. Nevertheless, over the past years, the progress achieved in the area of AI, as well as the commercial availability of Advanced Driver Assistance Systems (ADAS), has brought us closer to the goal of full driving autonomy~\cite{c1,c2}.

The Society of Automotive Engineers (SAE) has defined standard J3016 for different autonomy levels \cite{c3}, which splits the concept of autonomous driving into 5 levels of automation. Levels 1 and 2 are represented by systems where the human driver is required to monitor the driving scene, whereas levels 3, 4 and 5, with 5 being fully autonomous, consider that the automated driving components are monitoring the environment.

An AV must be able to sense its own surroundings and form an environment model consisting of moving and stationary objects~\cite{c4}. Afterwards it uses this information in order to learn long term driving strategies. These driving policies govern the vehicle’s motion~\cite{c5} and automatically output control signals for steering wheel, throttle and brake. At the highest level, the vehicle`s decision-making system has to select an optimal route from the current position to the destination~\cite{c6}.

The main reason behind the human ability to drive cars is our capability to understand the driving environment, or driving context. In the followings, we will refer to the driving context as the scene and we will define it as linked patterns of objects. In this paper, we introduce AIBA (\textit{AI Behavior Arbitration}), an algorithm designed to arbitrate between different driving strategies, or vehicle behaviors, based on AIBA’s understanding of the relations between the scene objects.

\begin{figure}
	\centering
	\captionsetup{justification=centering}
	\begin{center}
		\includegraphics[scale=1.23, width=\textwidth]{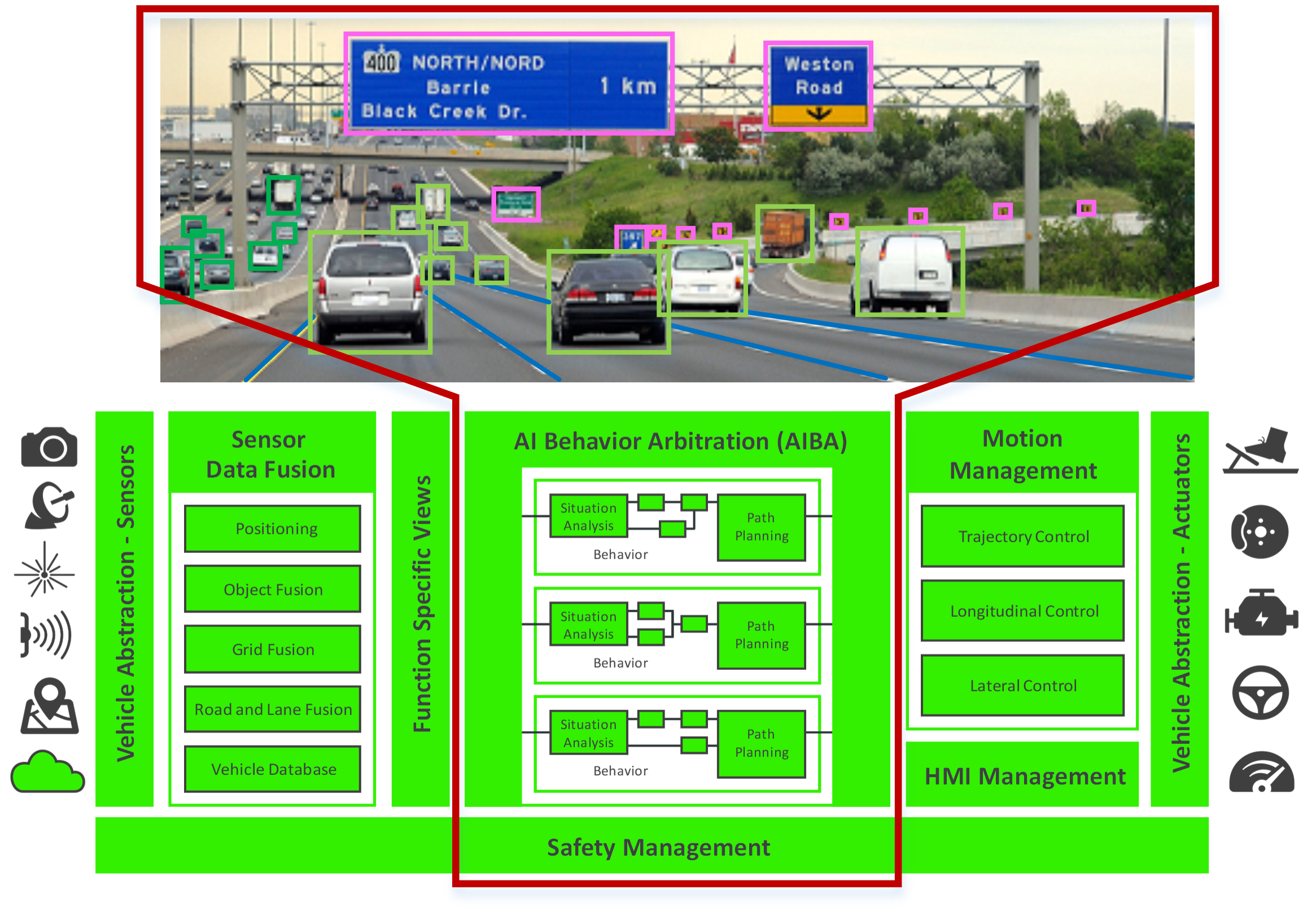}
		\caption{\textbf{Behavior Arbitration using AIBA} in the EB robinos autonomous driving framework}
        \label{fig:behavior_arbitration_aiba}
	\end{center}
	\vspace{-1.8em}
\end{figure}
 
A driving scene consists of objects such as lanes, sidewalks, cars, pedestrians, bicycles, traffic signs, etc., all of them being connected to each other in a particular way (e.g. a traffic sign displays information for a driver). An example of such a driving scene, processed by AIBA within EB robinos, is depicted in Figure 1. EB robinos is a functional software architecture from Elektrobit Automotive GmbH, with open interfaces and software modules that manages the complexity of autonomous driving. The EB robinos reference architecture integrates the components following the \textit{sense, plan, act} decomposition paradigm. Moreover, it also makes use of AI technology within its software modules in order to cope with the highly unstructured real-world driving environment.

Driving behavior prediction is a key part of an AV’s decision-making process. The task of identifying the future behavior of the scene’s objects is not a trivial one, since this type of information cannot be directly measured or communicated, being considered latent information~\cite{c7}. To be able to perform behavior prediction for the surrounding vehicles, an AV can use mathematical models which consider the variation in the objects’ movement and describe the driving scenario from the view point of the ego-car.

Such kind of models use several types of information, i.e. vehicle kinematics, the relationship between the ego-car and the surrounding entities, the interactions with other vehicles and a-priori knowledge. Vehicle kinematics and the relations with road entities were considered by almost all existing studies~\cite{c8}. For example, Lefevre et al. used the Time to Line Crossing (TTLC) to predict whether the vehicle will depart from the current lane or not~\cite{c9}.

Usually, most models for behavior arbitration are tailored for one specific scenario. Nevertheless, AVs must drive through dynamically changing environments in which a diversity of scenarios occur over time~\cite{c10}. Multiple scenario-specific models activate a corresponding model according to the characteristics of the scenario. In~\cite{c11}, the authors of this paper have proposed a three level Milcon type architecture. The second level, called the tactical level, includes a behavior collection and a manager program which trigger the appropriate AV behavior.

In this paper, we introduce AIBA, a system for behavior arbitration in AVs, which constructs a description and understanding model of the driving scene. Our idea is to model the human driver (HDr) understanding process of the driving scene, in order to achieve an optimal behavior arbitration solution. We describe the mechanism of the HDr thinking and transpose it to an approximate model. 

The rest of the paper is organized as follows. The concept for describing the driving scene is given in Section II, followed by the understanding and modelling presented in Section III. The experiments showcasing AIBA are detailed in Section IV. Finally, conclusions are stated in Section V.

\section{Driving Scene Description}

The driving scene description is given from a human driver's perspective, and it formulates properties derived from the definitions of classes, subclasses and objects which represent the core of an abstraction model, based on the authors' previous work~\cite{c12}. The main idea behind AIBA is to model, or formalize, the HDr understanding process and afterwards transform it into a formal model for behavior arbitration in AV.

\subsection{Driving Scene Analysis}

A human driver is able to perceive the scene's objects and observe them. This means that the HDr identifies the concepts and the different properties of the objects. In the end the driver can describe the scene in Natural Language (NL), as for the traffic example in Fig. 1: “the car is near exit E1; the main road continues straight and will reach the location L1, L2, L3; the traffic has a medium density and takes place under normal circumstances; the car is in the 3rd lane in safe vicinity to other cars". Apparently, this message eludes a lot of information, but if it is shared with other human drivers, it proves to be a piece of very important and comprehensive knowledge. 

The different scene objects are linked between them, as illustrated in the scene network from Fig. 2, built for the traffic scene shown in Fig. 1. Here, intuition assumes that the most important links are those established with the front car, or the main road (bold red lines in the image), while the less important links are those between the ego-car and the buildings (thin red lines in the image). In Fig. 2, each class of objects is linked with a rectangle. The symbols used are: \textit{AC} for the ego-vehicle autonomous car, \textit{Bld} for the buildings (i.e. side barriers), \textit{C1, C2, ... , Cn} for the other traffic cars, \textit{M, R, FR} for lane identification, and \textit{RS} for the road traffic signs.

\begin{figure}
	\centering
	\captionsetup{justification=justified}
	\begin{center}
		\includegraphics[scale=0.45]{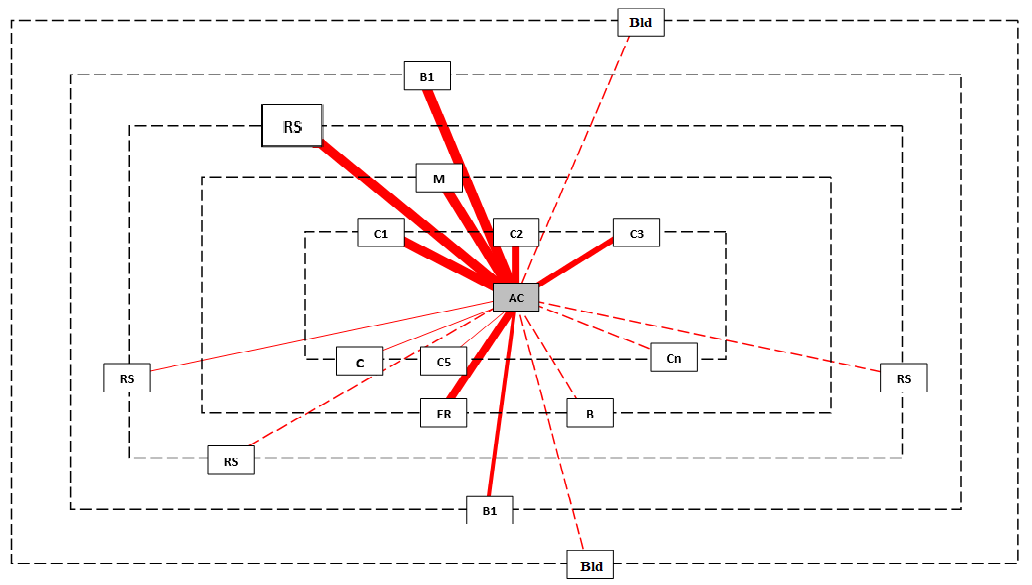}
		\caption{\textbf{Driving scene network} representing links between different objects, as imagined by a human driver. Bold red lines suggest a higher link importance level, whereas thin red lines point to a lower importance level.}
        \label{fig:driving_scene_net}
	\end{center}
	\vspace{-1.8em}
\end{figure}

The links definition is a first step in the knowledge process, which means it establishes the subjects of interest and also the importance level of each subject. The HDr scene understanding synthesis contains the following steps: identifying the link between the objects, allocating models to each link, running the models and afterwards finding a strategy to act. In fact, the HDr creates an implicit system and simulates it.

Two types of links can be observed: internal links, considered to be the set of links between the observer (the HDr) and all the observed objects, and external links, represented by the set of links between the objects present in the scene.  From the HDr point of view, the links have different meanings and importance, or significance.  Specifically, a human driver knows the traffic rules and how to get to his destination. These rules will determine traffic priorities, thus making the HDr to attach a greater importance to those links which are more important to his strategy.

In the next step, a description is established for each link. During the driving, the HDr adapts, or refines, the mentioned description by observation. The driver, by using the importance of the links, simulates a scene description. If we analyze the aim of driving scene understanding, we will observe that its origins are the stability in time and space. More precisely, the human driver has a driving task which can be accomplished if the possibility of locomotion is preserved in the current position and during a specific time. Intuitively, the stability is related to the objects in the scene and can be reached by understanding the scene. This understanding does not solve the driving problem, but offers the information upon which the human driver decides to act on a certain behavior. 

\subsection{Description proposal}

Several concepts can be proposed at this point of analysis:

\renewcommand{\labelitemi}{$\textasteriskcentered $}
\begin{itemize}
	\item The understanding is related with the concept of intelligence which is defined by Piaget as “the human ability to obtain stability in time and space”~\cite{c13}. This means that the links and the models (behaviors) are correlated to the HDr intelligence.
	\item The driving task has multiple attributes: determine a trajectory from start to end, ensure the comfort, minimize the energy consumption etc.,all which are time dependent. The task trajectory must be accomplished in safe conditions and should be included in the stability concept.
	\item The driving scene is a complex environment, many entities being linked. In order to handle this complexity, the HDr establishes a network representing a holistic interpretation and priority interactions for each link of the network
	\item Anticipating, the AD system is equipped with a sensor network, giving highly qualitative information about the scene. 
\end{itemize}

Resuming, the previous analysis of the driving scene understanding can be described like a process which consists of the following steps:

\begin{enumerate}
	\item Perceiving the objects (cars, traffic signs, lanes, pedestrians, etc.) and obtaining the object properties (the pedestrian intention is to cross the road).
	\item Defining the links (the scene network) between the objects and their importance (i.e. the car in front is important, the pedestrian which intends to cross the road is very important, etc.).
	\item Adding models to the mentioned links.
	\item Simulating the model of the most significant links, and proposing driving behaviors which will be verified with the other significant links until the appropriate behavior (which allows the driving task) is found.
\end{enumerate}

\section{Driving Scene Modelling}

Entities like objects, links, or networks, which have been introduced in the previous section, have correspondences in the modelling process. Our intention is to approximate the HD understanding process through a formal representation. More precisely, this assignment mimics an input/output process: using a perceived scene, the AIBA model must output a description which offers all the information needed within the AD system to arbitrate the driving behavior.

\begin{figure*}[thpb]
	\centering
	\captionsetup{justification=centering}
	\begin{center}
		\includegraphics[scale=0.37]{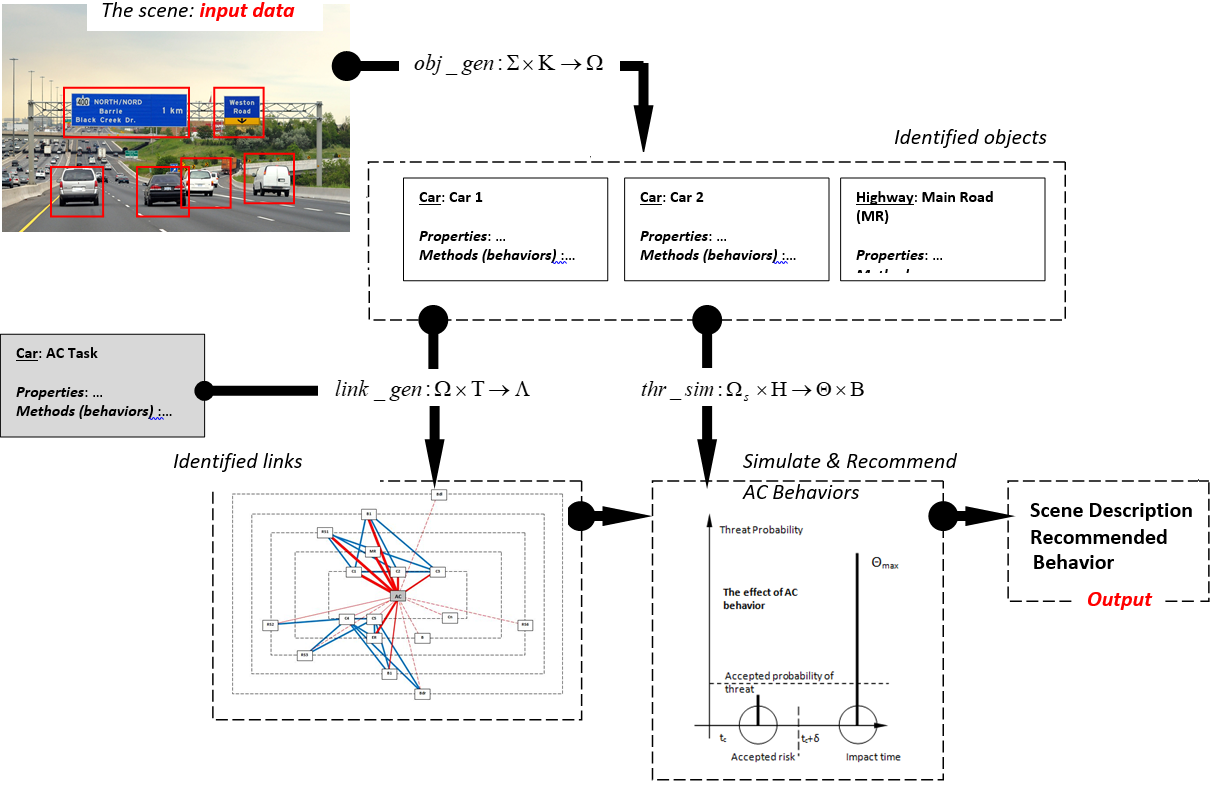}
		\caption{\textbf{The block diagram} of AIBA, overall picture.}
        \label{fig:block_diagram}
	\end{center}
	\vspace{-1.8em}
\end{figure*}

The block diagram representation of AIBA is illustrated in Fig.~\ref{fig:block_diagram}, where the information flow emulates the human driver scene understanding phenomenon.

Within AIBA, the first action is to transform the scene in to a collection of objects. This operation is accomplished by the following generative function:

\begin{equation}
	obj\_gen: \Sigma \times K \rightarrow \Omega
	\label{eq:generator_function}
\end{equation}

\noindent where $\Sigma$ is the perceived scene; $K$ is the set of known object classes and $\Omega$ is the set of objects.

Having an initial collection of classes, the generator function from Eq.~\ref{eq:generator_function} transforms the scene entities into a collection of objects. The following object classes are taken into consideration: traffic participants, pedestrians and buildings. The set of classes are a priori set within AIBA.

The complexity of this process, even for the set of road classes, is seen in many types of roads which exist around the world. In order to reduce the scene's complexity, we split them in two major classes, static objects (lanes, traffic signs, buildings, etc.) and dynamic objects (cars, pedestrians, etc.).

Two kinds of definitions are here significant: the generic definition where the proximity and the properties are mentioned, and the extensive definition where the definition of the object (or a picture of it) is indicted or shown (similar to Fig.~\ref{fig:behavior_arbitration_aiba}). This observation enables us to associate image recognition methods (which correspond to the extensive definition) with a generic definition collection.

Eq.~\ref{eq:generator_function} can be generalized when measurements (speed of the cars, distance between cars, size of traffic signs etc.) are associated with the scene description:

\begin{equation}
	obj\_gen: \Sigma \times K \times M \rightarrow \Omega,
\end{equation}

\noindent where $M$ is the set of measurements. The generated objects $\Omega$ contain a structure of class specific properties and methods. The methods reflect the possible interactions of the objects with the ego-car.

The second step in AIBA's workflow defines the links between a HDr and an object, as well as between the objects themselves, respectively:

\begin{equation}
	link\_gen: \Omega \times T \rightarrow \Lambda,
	\label{eq:links}
\end{equation}

\noindent where $T$ is the set of task trajectory properties and $\Lambda$ is the set of links' significance.

Because the links are computed in term of stability around a particular point of the task trajectory, the significance is correlated with specific threats. Driving a car is subject to implicit negotiation based on traffic rules. The most important links are those related to agents which, according to these rules, have priority. Eq.~\ref{eq:links} can be imagined as an expert system which will analyse all these links from the mentioned point of view, while outputting different marks:

\begin{equation}
	obj\_gen: \Sigma \times K \times M \times T \rightarrow \Omega \times \Lambda.
\end{equation}

Each recognized object provides methods which refer to the ego-car - object interactions. Information about possible threats on the task stability can be obtained by simulating the behaviors, and also how to select the appropriate driving behavior for avoiding these threats:

\begin{equation}
	thr\_sim: \Omega_S \times H \rightarrow \Theta \times B,
\end{equation}

\begin{equation}
	\Omega_S = \left\{ O_i | O_i \in \Omega; S_{O_i, O_j} \geq S_{min} \right\}
\end{equation}

\begin{equation}
	H = \begin{bmatrix} t_c && t_c + \delta \end{bmatrix}
\end{equation}

\noindent where $\Omega_S$ is the set of important objects $O_i$, which are linked with other objects $O_j$ with a significance $S_{O_i, O_j} \in \Lambda$, greater than a minimum (a priori imposed) significance $S_{min}$. $H$ is the time horizon, $t_c$ is the current time, $\delta$ is the simulation time, $\Theta$ is the set of the threat levels and $B$ is the set of recommended behaviors for the ego-car.

$thr\_sim$ will simulate, for each important object $O_i$, a collection of behaviors $O_i\_M_{i,k} (P, H)$ and a predicted threat level $\Theta_{i,k}$:

\begin{equation}
	O_i\_M_{i,k} (P, H) = 
		\begin{bmatrix}
			\Theta_{i,k} \\
			\beta_{i,k}
		\end{bmatrix}
\end{equation}

\noindent where $P$ is the set of object properties and $\beta_{i,k}$ is the behavior of the ego-car which will eliminate the threat. The mentioned models can be analytical functions~\cite{c2}, fuzzy engines~\cite{c14}, or even neural networks~\cite{c15}.

In order to solve all links' threats, several strategies can be chosen. If we adopt the HDr understanding description from the previous section, the behavior which solves the maximum threat is simulated for the other threats, obtaining the optimal behavior of the ego-car $\Theta_{max} = \Theta_{i^*, k^*}$:

\begin{equation}
	(i^*, k^*) = \arg \max_{i,k} \Theta_{i,k}
\end{equation}

\noindent where $\Theta_{i,k}$ are the threat levels. 

The last step in AIBA's modelling system is the transformation of the optimal behavior $\Theta_{max}$ into a natural language explanation:

\begin{equation}
	dsc: \Lambda \times B \rightarrow \Delta
\end{equation}

\begin{equation}
	\Delta = \Omega_S \times E
\end{equation}

\begin{equation}
	E = e_1 \times e_2 \times ... \times e_{n_S}
\end{equation}

\noindent where $\Delta$ is the set of descriptions, $e_i$ is the explanation of the significance associated to an object and $n_S$ is the number of significant objects.

\section{Experiments}

Our experiments were conducted in our own autonomous vehicles prototyping simulator GridSim~\cite{c17} and in Virtual Test Drive(VTD).

GridSim is a two dimensional birds' eye view autonomous driving simulator engine, which uses a car-like robot architecture to generate occupancy grids from simulated sensors. It allows for multiple scenarios to be easily represented and loaded into the simulator as backgrounds, while the kinematic engine itself is based on the single-track kinematic model.

In such a model, a no-slip assumption for the wheels on the driving surface is considered. The vehicle obeys the "non-holonomic" assumption, expressed as a differential constraint on the motion of the car. The non-holonomic constraint restricts the vehicle from making lateral displacements, without simultaneously moving forward.

\begin{figure}
	\centering
	\captionsetup{justification=justified}
	\begin{center}
		\includegraphics[scale=0.27]{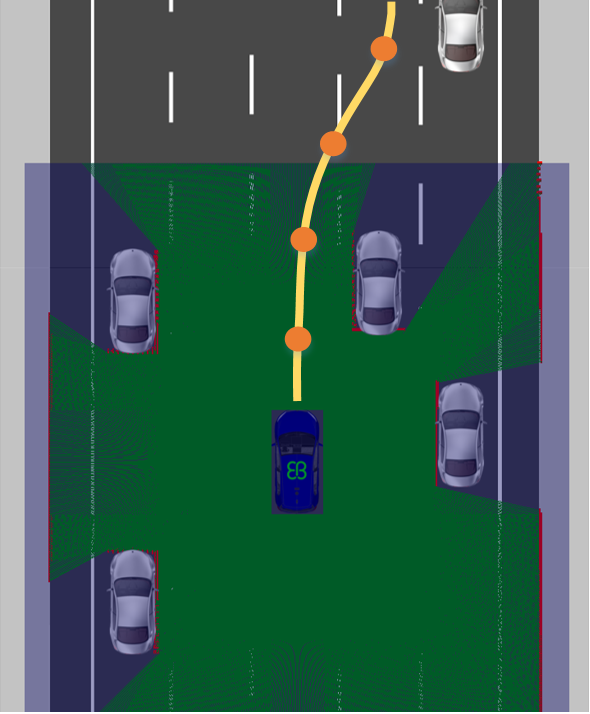}
		\caption{\textbf{GridSim} example of a simulation scenario. The green area is a simulated occupancy grid, used to enhance the measurement of the properties of objects in the scene.}
        \label{fig:gridsim_example}
	\end{center}
	\vspace{-1.8em}
\end{figure}

Occupancy grids are often used for environment perception and navigation, applications which require techniques for data fusion and obstacles avoidance. We used such a representation in the previous work for driving context classification~\cite{c18}. We assume that the driving area should coincide with free space, while non-drivable areas may be represented by other traffic participants (dynamic obstacles), road boundaries, buildings, or other static obstacles. The virtual traffic participants inside GridSim are generated by sampling their trajectory from a uniform distribution which describes their steering angle's rate of change, as well as their longitudinal velocity. GridSim determines the sensor's freespace and occupied areas, where the participants are considered as obstacles. This representation shows free-space and occupied areas in a bird’s eye perspective. An example of such a representation can be seen in Fig.~\ref{fig:gridsim_example}.

VTD is a complete tool-chain for driving simulation applications. The tool-kit is used for the creation, configuration, presentation and evaluation of virtual environments in the scope of based simulations. It is used for the development of ADAS and automated driving systems as well as the core for training simulators. It covers the full range from the generation of 3D content to the simulation of complex traffic scenarios and, finally, to the simulation of either simplified or physically driven sensors.

\begin{figure}
	\centering
	\captionsetup{justification=centering}
	\begin{center}
		\includegraphics[scale=0.26]{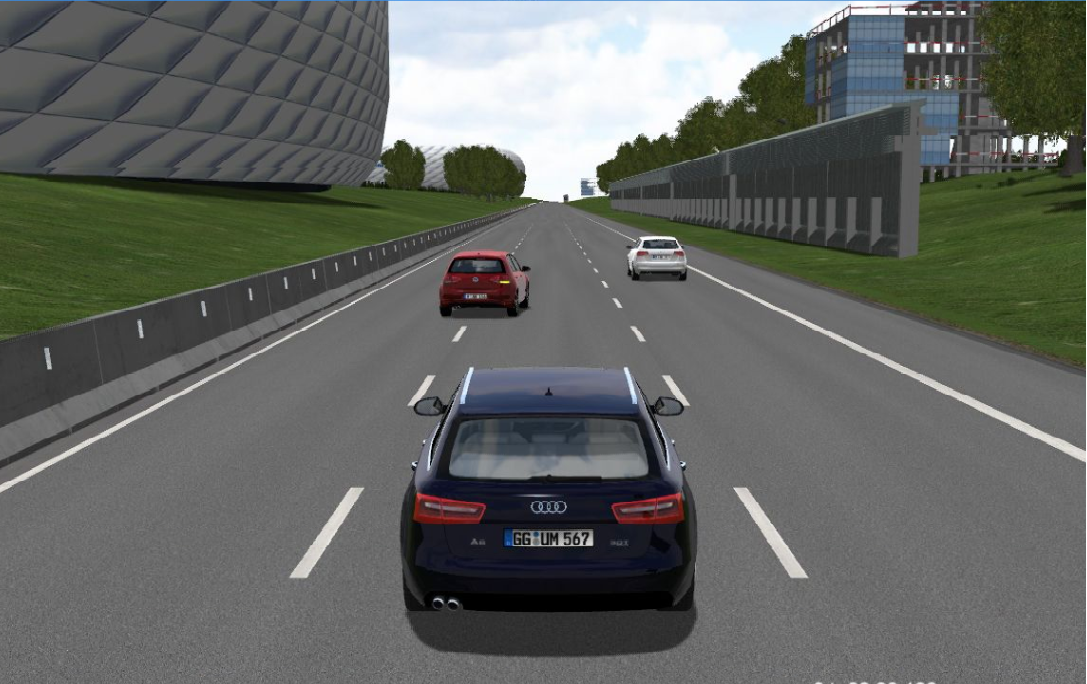}
		\caption{\textbf{VTD - Virtual Test Drive} example of a simulation scenario.}
        \label{fig:vtd_example}
	\end{center}
	\vspace{-1.8em}
\end{figure}

\begin{figure}
	\centering
	\captionsetup{justification=centering}
	\begin{center}
		\includegraphics[scale=0.48]{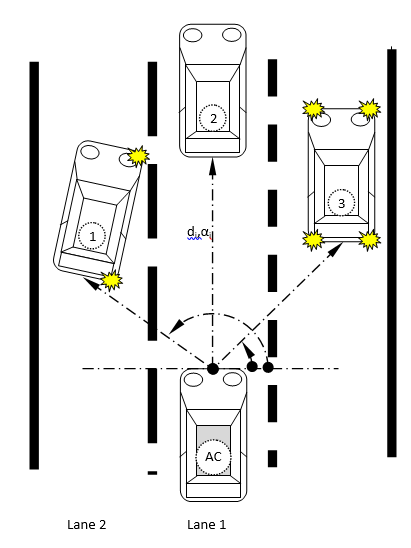}
		\caption{\textbf{Sketch} of the simulation scenario}
        \label{fig:simulation_scenario}
	\end{center}
	\vspace{-1.8em}
\end{figure}

For better understandability, we explain the behavior arbitration steps on a proposed simulation scenario. The sketch of the such a scenario can be found in Fig.~\ref{fig:simulation_scenario}. The car is driving in the middle lane, behind a car in the same lane. A car from the left side signals its intention to change the lane to the right, in front of the ego-vehicle.

In the first step the ego-vehicle identifies the three objects from the scene and measures the properties, also computing the possible impact times:

\renewcommand{\labelitemi}{$\bullet$}
\begin{itemize}
\item Car 1: in lane 3 (left, near), at distance \textit{d1}, with speed \textit{v1}, changing the lane
\item Car 2: in lane 2 (in front, far), at distance \textit{d2}, with speed \textit{v2}, following the 2nd lane
\item Car 3: in lane 1 (right, near), at distance \textit{d3}, with speed \textit{v3}, intending to stop
\end{itemize}

\begin{table}[h]
	\caption{Table with resulted impact times. The highlighted values are the ones that pose threats to the ego-vehicle.}
	\label{table_1}
	\begin{center}
		\renewcommand{\arraystretch}{1.2}
		\begin{tabular}{ |c||c|c|c|c| } 
 			\hline
 			\multirow{2}{*}{Car} & LaneFollow & LaneChangeRight & LaneChangeLeft & Stop \\ 
 			\hhline{|~|-|-|-|-|}
 			& \multicolumn{4}{|c|}{Impact Time} \\
 			\hline\hline
 			\multirow{2}{*}{1} & 0,2 & \cellcolor{yellow!90}0,59  & 0,01 & 0,2 \\ 
 			\hhline{|~|-|-|-|-|}
 			& $\infty$ & \cellcolor{yellow!90}20 & $\infty$ & $\infty$ \\ 
 			\hline
 			\multirow{2}{*}{2} & \cellcolor{yellow!75}0,6 & 0,1 & 0,1 & \cellcolor{yellow!90}0,2 \\
 			\hhline{|~|-|-|-|-|}
 			& \cellcolor{yellow!90}25 & $\infty$ & $\infty$ &\cellcolor{yellow!90} 10 \\ 
 			\hline
 			\multirow{2}{*}{3} & 0,2 & 0,2 & \cellcolor{yellow!90}0,1 & 0,5 \\ 
 			\hhline{|~|-|-|~|-|}
 			& $\infty$ & $\infty$ & \cellcolor{yellow!90}30 & $\infty$ \\
 			\hline 
		\end{tabular}
	\end{center}
	\vspace{-1.8em}
\end{table}

In the second step the ego-vehicle identifies links level:

\begin{itemize}
\item Car 1 (left, near) $\lambda_1$ = 0.3
\item Car 2 (in front, far) $\lambda_2$ = 0.6
\item Car 3 (right, near) $\lambda_3$ = 0.1
\end{itemize}

In the third step the threat time is computed, and compared to the maximum accepted level:

\begin{figure}
	\centering
	\captionsetup{justification=centering}
	\begin{center}
		\includegraphics[scale=0.08]{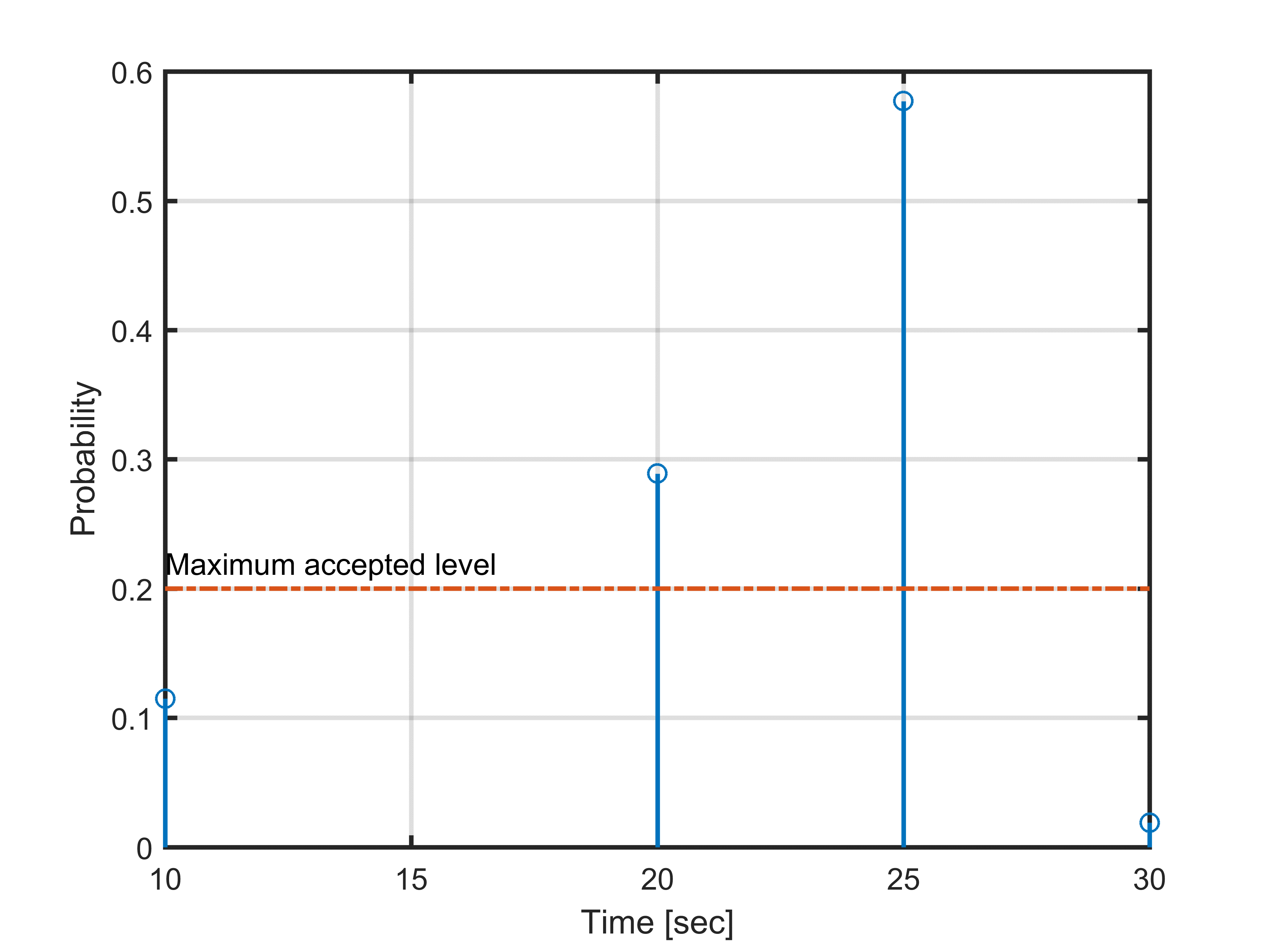}
		\caption{Maximum accepted level of all threats, on a time axis. }
        \label{fig:graph_eval}
	\end{center}
	\vspace{-1.8em}
\end{figure}

In the fourth and final step, the behavior of the ego-vehicle which solves the most probable threat is reducing speed. This is also simulated for the other threats which are above the maximum accepted level, and since it also solves the lane change to the right of the first car, it is chosen as the optimal behavior of the ego-car.

\section{CONCLUSION}

In summary, this paper proposes a novel approach to address the task of behavior arbitration. The method, coined AIBA, aims at formalizing and learning human-like driving behaviors from the description, understanding and modelling of a traffic scenario. The main advantage of the formal representation is that the functional safety of the automotive system can be analytically inferred, as opposed to a neural network black-box, for example. 

As future work, we would like to extend the proposed method to more granular driving scenarios, such as driving in an indoor parking lot, or driving on uncharted roads. Additionally, we would like to add more objects in the scene, and to test the deployment on an embedded device.


\section*{ACKNOWLEDGMENTS}

We hereby acknowledge Elektrobit Automotive, Széchenyi István University, and the TAMOP - 4.2.2.C-11/1/KONV-2012-0012 Project "Smarter Transport - IT for co-operative transport system" for providing the infrastructure and for support during research.


\end{document}